\documentclass[runningheads]{paper1557}
\usepackage[T1]{fontenc}
\usepackage{float}
\usepackage{multirow}
\usepackage{booktabs}
\usepackage{amsmath}
\usepackage{amssymb}
\usepackage{makecell}
\usepackage{marvosym}

\usepackage{graphicx}
\usepackage{xcolor}
\usepackage{subfiles}

\makeatletter

\newcommand{\Rmnum}[1]{\expandafter\@slowromancap\romannumeral #1@}
\makeatother
\newcommand{\para}[1]{\vspace{.05in}\noindent\textbf{#1}}

\def\ie{\emph{i.e.}}

\def\etal{{\em et al.}}

\begin{document}
\title{Joint Prediction of Meningioma Grade and Brain Invasion via Task-Aware Contrastive Learning}
%
%

\author{Tianling Liu\inst{1} \and
Wennan Liu\inst{2} \and
Lequan Yu\inst{3}\and
Liang Wan\inst{1}\textsuperscript{(\Letter)} \and
Tong Han\inst{4} \and
Lei Zhu\inst{5,6}
}


%
\authorrunning{T. Liu~\etal}
%
\institute{College of Intelligence and Computing, Tianjin University, Tianjin, China
\email{liu\_dling@tju.edu.cn, lwan@tju.edu.cn} \and
Medical College of Tianjin University, Tianjin, China \and
The University of Hong Kong, Hong Kong, China \and
Brain Medical Center of Tianjin University, Huanhu Hospital, Tianjin, China \and
The Hong Kong University of Science and Technology (Guangzhou), Guangzhou, China \and
The Hong Kong University of Science and Technology, Hong Kong, China
}
\titlerunning{Joint Prediction of Meningioma Grade and Brain Invasion}
\maketitle              
\begin{abstract}

Preoperative and noninvasive prediction of the meningioma grade is important in clinical practice, as it directly influences the clinical decision making.
What's more, brain invasion in meningioma (\ie, the presence of tumor tissue within the adjacent brain tissue) is an independent criterion for the grading of meningioma and influences the treatment strategy.
Although efforts have been reported to address these two tasks, most of them rely on hand-crafted features and there is no attempt to exploit the two prediction tasks simultaneously.
In this paper, we propose a novel task-aware contrastive learning algorithm to jointly predict meningioma grade and brain invasion from multi-modal MRIs.
Based on the basic multi-task learning framework, our key idea is to adopt contrastive learning strategy to disentangle the image features into task-specific features and task-common features, and explicitly leverage their inherent connections to improve feature representation for the two prediction tasks.
In this retrospective study, an MRI dataset was collected, for which 800 patients (containing 148 high-grade, 62 invasion) were diagnosed with meningioma by pathological analysis. 
Experimental results show that the proposed algorithm outperforms alternative multi-task learning methods, achieving AUCs of $0.8870$ and $0.9787$ for the prediction of meningioma grade and brain invasion, respectively. The code is available at https://github.com/IsDling/predictTCL.
 
\end{abstract}

\section{Introduction}

\begin{figure}[t]
    \centering
    \includegraphics[width=\textwidth]{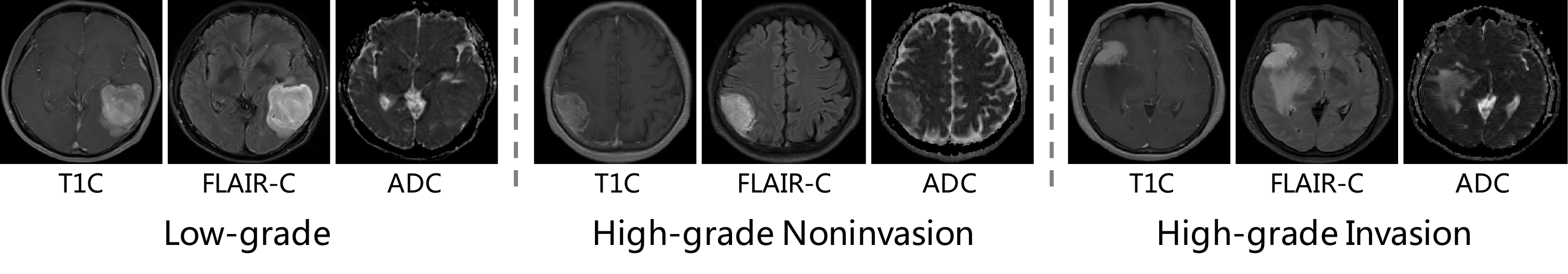}
    \vspace{-7mm}
    \caption{MRI examples for low-grade meningiomas, high-grade meningiomas without/with brain invasion.}
    \label{MRI_exhibit}
\end{figure}

Meningiomas are the most common primary intracranial tumors in adults, comprising 38.3\% of central nervous system tumors~\cite{Ostrom2020CBTRUS}.
According to the World Health Organization (WHO), meningiomas can be subdivided into three grades, \ie, grade~\Rmnum{1} (80\%), grade~\Rmnum{2} (18\%) and grade~\Rmnum{3} (2\%).  
Grade~\Rmnum{1} represents low-grade meningiomas, grade~\Rmnum{2} and grade~\Rmnum{3} are high-grade meningiomas. 
The WHO grading is an important factor in determining treatment options and overall prognosis for meningiomas.
Specifically, low-grade meningiomas can be treated with surgery or beam radiation, and rarely recur after resection, while high-grade meningiomas should be treated with both means, and subjected to universal recurrence for grade III and 20–75\% recurrence rates for grade II~\cite{Champeaux2017Recurrence,huang2019imaging}. 
In another aspect, brain invasion is taken as a stand-alone pathological criterion for distinguishing between grade~\Rmnum{1} and grade~\Rmnum{2} in 2016 WHO classification~\cite{louis20162016}. 
Recent researches also uncover a link between brain invasion and increased risks of tumor progression, disease recurrence and poor prognosis~\cite{nowosielski2017diagnostic,brokinkel2017brain,li2021clinical}. 

In real clinical diagnosis, pathological analysis provides the gold standard for the determination of brain invasion and meningioma grading~\cite{louis20072007,louis20162016}.
However, to conduct pathological analysis, clinicians are required to sample tissues from the core and surrounding areas of the tumor during \textit{invasive} resection or biopsy, while some important treatment decision has been made without the knowledge of brain invasion and meningioma grading.
Moreover, the accuracy of brain invasion determination heavily depends on the clinician's experience~\cite{zhang2020radiomics}. 
If the brain tissue samples do not fall in the invasion area, there is a risk that the patient will be misdiagnosed and his prognosis can be affected. 
Given these practical concerns, accurate \textit{preoperative} and \textit{non-invasive} assessment of meningioma grade and brain invasion is clinically essential to facilitate treatment decisions.

Many recent studies in the clinical field are devoted to predicting the grade of meningioma or identification of brain invasion by analyzing brain MRIs~\cite{hale2018machine,park2019radiomics,zhu2019deep,han2021meningiomas,zhang2021deep,KANDEMIRLI2020106205,zhang2020radiomics}. 
Hale~\etal~extracted radiographic features by traditional statistical methods and verified that machine learning-based methods can help to predict meningioma grade~\cite{hale2018machine}. 
Later on, several followed-up studies are reported to extract different hand-crafted features or CNN features and then use machine learning-based classifiers, such as SVM or random forest, to predict the meningioma grade~\cite{park2019radiomics,zhu2019deep,han2021meningiomas}, or for the determination of brain invasion~\cite{KANDEMIRLI2020106205,zhang2020radiomics}.
Moreover, Zhang~\etal~applied the widely-used deep-learning network, ResNet, to predict the meningioma grade~\cite{zhang2021deep}. 
Overall, most of the previous studies extracted hand-crafted features and conducted the two tasks, without exploring the powerful feature representation learning of neural networks.  
Furthermore, to the best of our knowledge, there is no previous study performing both tasks simultaneously, although there exist clinical connections between them. 

\begin{figure}[!t]
    \centering
    \includegraphics[width=\textwidth]{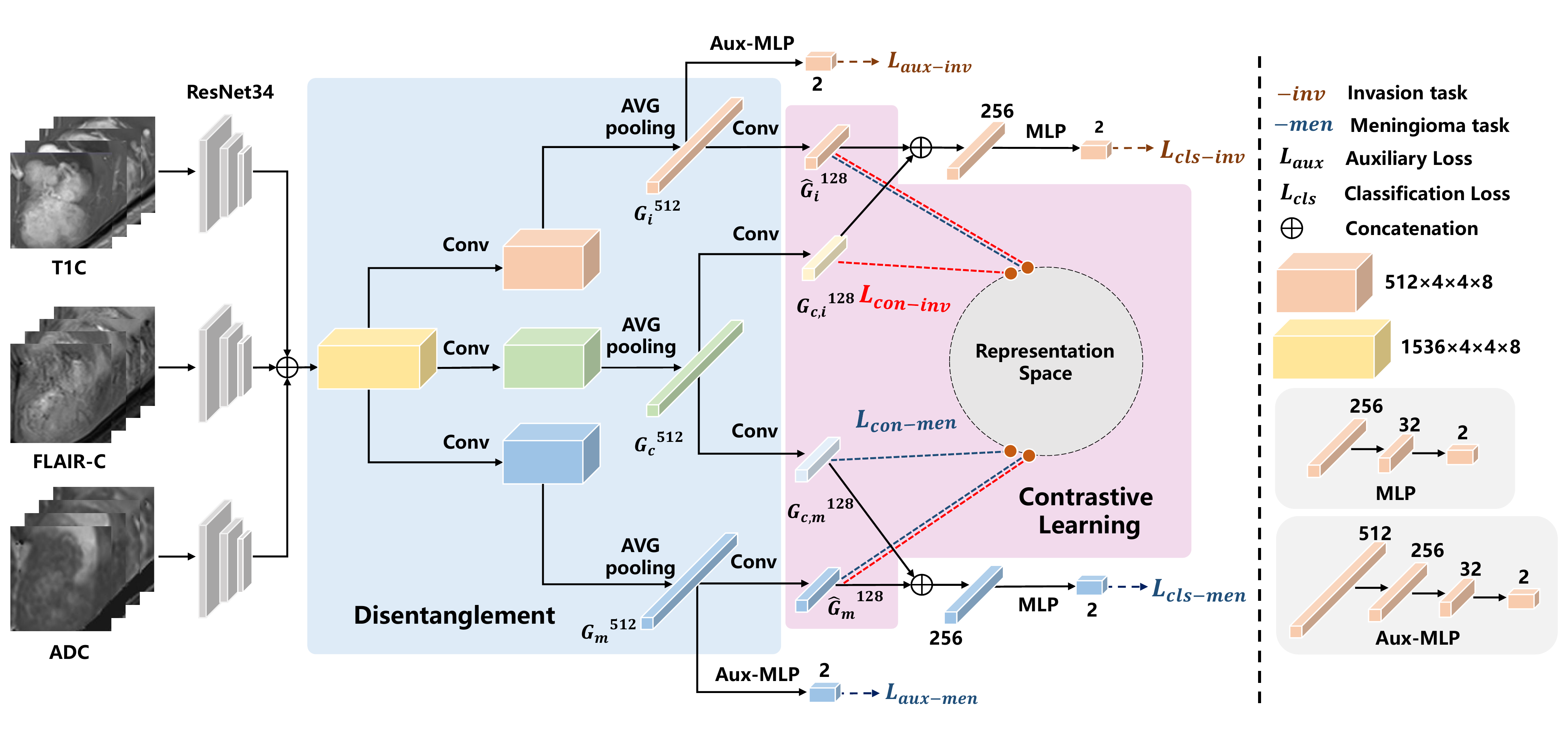}
    \vspace{-7mm}
    \caption{The overview of the proposed multi-task learning framework. We disentangle the fused feature maps into two task-specific features and one task-common feature.
    A task-aware contrastive learning strategy is devised to ensure that task-common features can better guide both tasks.
    }
    \label{fig:proposed_overview}
    \vspace{-4mm}
\end{figure}

In this paper, as a preliminary exploration, we develop a novel multi-task learning algorithm to make joint prediction of brain invasion and meningioma grade from multi-modal MRIs, including  post-contrast T1-weighted (T1C), constrat T2 fluid attenuation inversion recovery (FLAIR-C), and apparent diffusion coefficient (ADC) images, following clinical practice (Fig.~\ref{MRI_exhibit}).
Note that the common multi-task learning strategy adopts a shared backbone and then learns task-specific features with multiple heads to perform separate  tasks~\cite{Doersch2017multitask}.
Existing methods are usually focused to tune the loss function to balance contributions between different tasks~\cite{Chen2018multitask,Kendall2018multitask}, which is still weak to ensure the feature representation ability.
We notice that multi-task methods assume that the tasks are related to each other, and considering tasks simultaneously can help to improve the feature representation ability. 
Hence, in our work, we propose a novel task-aware contrastive learning strategy, by respecting both the coherence between tasks and the distinctness of each individual task. 
Our approach disentangles the multi-modal MRI features into \textit{task-specific} features, which are sensitive to a certain classification task, as well as \textit{task-common} features that are helpful to both prediction tasks.
Then we align the task-common features to each task, and enforce the feature emeddings contributing to the same task to be more similar than the embeddings contributing to different tasks, via contrastive losses. 
%
%

In summary, our contributions are three folds:
(1) This is the first study to simultaneously predict meningioma grade and identify brain invasion.
(2) We propose a simple yet effective task-aware contrastive learning strategy, which derives task-common features in addition to task-specific features from the shared feature encoder, and takes task-common features as a guidance to improve the prediction ability for both tasks; 
(3) Experiments on our own collected dataset demonstrate that the proposed approach can  accurately predict meningioma grade and brain invasion, and outperforms alternative methods effectively.

\section{Method}
Fig.~\ref{fig:proposed_overview} demonstrates the architecture of the proposed multi-task framework. 
Given multi-modal brain images, we apply multiple backbones to extract the feature maps from MRIs respectively, and here ResNet34 is adopted as the backbone.
The fused features from multiple backbones contain rich information for performing multiple tasks.
We then disentangle the fused feature maps to task-specific features and task-common features,
and a new task-aware contrastive learning strategy is leveraged to exploit the “comparative” relation between task-specific features and task-common features. 

\subsection{Feature Disentanglement}
As demonstrated by~\cite{chartsias2019disentangled,chartsias2020disentangle}, the disentangled representation, which decouples the entangled features into task-specific features that are easier to predict, is beneficial to the realization of multi-task learning. 
Like most multi-task applications, our goal is to find the most discriminative features for each prediction task. 
Although the general practice is to disentangle the fused features into multiple task-specific features, we consider that there are features that are simultaneously effective for multiple tasks, since multi-task learning assumes that the tasks are related to each other.
In our work, we disentangle the fused features into the task-specific feature for invasion identification task (denoted as $G_{i}$), the task-specific feature for meningioma grading task ($G_{m}$), and the task-common feature ($G_{c}$). Here, we adopt a convolution layer and an average pooling to realize feature disentanglement, which is formulated as follows,
\begin{equation}
\begin{split}
     G_{k} = AP(Conv(Concat(F_{T},F_{F},F{_A}))), k=i,m,c,
\end{split}
\end{equation}
where \{$F_{T}, F_{F}, F_{A} \in \mathbb{R}^{C \times h \times w \times d}$\} denote the feature maps extracted from T1C, Flair-C and ADC MRIs respectively, and their sizes are all $512\times4\times4\times8$. 
$Concat(\cdot)$ means channel concatenation. 
$Conv(\cdot)$ means $2\times2$ convolution operation, $AP(\cdot)$ is average pooling. After flattening, the resultant disentangled features $\{G_{i}, G_{m}, G_{c}\}$  are feature vectors with size of $512$.
Note that to realize feature disentanglement, we will rely on the task-aware contrastive learning as well as auxiliary classification branches,  discussed in the following subsections.

\subsection{Task-Aware Contrastive Learning}
The proposed task-aware contrastive learning strategy has two functions, one is to help feature disentanglement, and the other is to improve the predictive ability of each task-specific features for the corresponding task by leveraging the relationship between task-common features and task-specific features.
Contrastive learning (CL) has shown great potential in the natural image field, and has been applied in the medical image filed in recent years. 
Its core idea is ``learn to compare'': given an anchor point, distinguish a similar (or positive) sample from a set of dissimilar (or negative) samples, in a projected embedding space~\cite{wang2021exploring}.
Most current CL researches are image-level and pixel-level. 
Image-level CL~\cite{henaff2020data,xie2020delving} takes multiple views of the same image as positive samples and different images as negative samples. Pixel-level CL~\cite{wang2021exploring,xie2021propagate} take pixels from the same class as positive samples and pixels from different classes as negative samples. 
Different from CL methods mentioned above, we propose a new task-aware contrastive learning strategy. 
As the task-common feature is supposed to be helpful for both tasks, we align it to each task and enforce the feature emeddings contributing to the same task to be more similar than the embeddings contributing to different tasks, via contrastive
losses.  
This simple strategy respects the coherence
between tasks and strengthens the distinctness of each individual task, thus improving the feature representation ability of the network.

Taking the invasion identification task as an example, we align the task-common feature $G_{c}$ to this task to get $G_{c,i}$ via a fully connected layer, yielding $G_{c,i} \in \mathbb{R}^{128}$.
Then $G_{c,i}$ is expected to be similar to the task-specific feature for invasion identification $G_{i}$, and not like the task-specific feature for meningioma grading $G_{m}$. 
To this end, we also transform two task-specific features into $\hat{G}_i$ and $\hat{G}_m$ via fully connected layers, respectively, where $\hat{G}_i,\hat{G}_m \in \mathbb{R}^{128}$. 
We define the task-aware contrastive loss for the invasion identification task as
\begin{equation}
    \mathcal{L}_{con-inv}=-log\frac{exp(sim(G_{c,i},\hat{G}_i)/\tau)}{exp(sim(G_{c,i},\hat{G}_i)/\tau)+exp(sim(G_{c,i},\hat{G}_m)/\tau)},
\end{equation}
where $sim(\cdot)$ denotes cosine similarity between two feature vectors; $\tau$ is the temperature parameter and set as 0.07 empirically.
Similarly, the task-aware contrastive loss for the meningioma grading task is defined as follows,
\begin{equation}
    \mathcal{L}_{con-men}=-log\frac{exp(sim(G_{c,m},\hat{G}_m)/\tau)}{exp(sim(G_{c,m},\hat{G}_m)/\tau)+exp(sim(G_{c,m},\hat{G}_i)/\tau)},
\end{equation}
where $G_{c,m} \in \mathbb{R}^{128}$ denotes the manipulated feature vector via aligning the task-common feature $G_{c}$ to meningioma grading.

\subsection{Overall Loss Functions and Training Strategy}
Besides to estimate the task-aware contrastive losses, the features for the same task are further concatenated and undergo a three-layer MLP (with dimensions of 256, 32 and 2), yielding the prediction loss for this task, i.e. $\mathcal{L}_{cls-inv}$ and $\mathcal{L}_{cls-men}$.
In order to promote the guiding ability of task-common features, we introduce auxiliary classification branches to the disentangled task-specific features; see Fig.~\ref{fig:proposed_overview}, which offers two auxiliary classification losses, i.e. $\mathcal{L}_{aux-inv}$ and $\mathcal{L}_{aux-men}$. 
Each auxiliary branch is a four-layer MLP with dimensions of 512, 256, 32 and 2.
In summary, our multi-task learning loss is defined as
\begin{equation}
\begin{split}
    \mathcal{L} = \mathcal{L}_{cls-inv}&+\mathcal{L}_{cls-men}+\alpha(\mathcal{L}_{con-inv}+\mathcal{L}_{con-men}) \\
    & +\beta(\mathcal{L}_{aux-inv}+\mathcal{L}_{aux-men}).
\end{split}
\end{equation}
We use cross entropy loss as the classification loss; $\alpha$ and $\beta$ are weights to balance the contribution of different losses, which are set as $1$ and $0.7$ empirically. 
Besides, to ensure the contrastive learning strategy work well, we add the contrastive learning loss after a period of training, empirically set as 30 epoches.

\section{Experiments}

\begin{table}[t]
\centering
\caption{Details of the dataset.}
\label{dataset_Details}
\setlength{\tabcolsep}{1.3mm}{
\small
\begin{tabular}{|c|c|cc|}
\hline
\multirow{2}{*}{Characteristics} & \multirow{2}{*}{Low grade} & \multicolumn{2}{c|}{High grade}                 \\ \cline{3-4} 
                                 &                            & \multicolumn{1}{c|}{invasion}    & noninvasion \\ \hline
Number                           & 652                        & \multicolumn{1}{c|}{62}          & 86          \\ \hline
Age(years$\pm$SD)                     & 56.75$\pm$10.7                 & \multicolumn{1}{c|}{55.77$\pm$11.97} & 55.52$\pm$12.46 \\ \hline
Male                             & 139                        & \multicolumn{1}{c|}{26}          & 39          \\ \hline
Female                           & 513                        & \multicolumn{1}{c|}{36}          & 47          \\ \hline
\end{tabular}}
\vspace{-0.3cm}
\end{table}

\para{Dataset and Preprocessing.} We collected an MRI dataset of meningiomas for patients with tumor resection between March 2016 and March 2021 in Brain Medical Center of Tianjin University, Tianjin Huanhu Hospital.
MRI scans were performed with four 3.0T MRI scanners (i.e., Skyra, Trio, Avanto, Prisma from Siemens).
Table~\ref{dataset_Details} presents details of the dataset, which contains 800 MRI volumes with a size of $256\times256\times24$ and $1mm$ spacing. 
Every MRI volume contains three modals, \ie, T1C (contrast-enhanced T1), FLAIR-C (contrast-enhanced T2 FLAIR) and ADC, and has two labels (\ie, grading and invasion classification labels) for each patient.

During experiments, due to the small amount of invasion samples, we use randomly drawn data division to alleviate overfitting.
Specifically, we randomly draw training and testing sets in three runs to ensure training and testing sets have similar distribution of low/high, invasion yes/no. 
Then there are 214 MRIs for training and the remaining MRIs as the testing dataset in each run. Specifically, the training dataset contains 44 invasion MRIs and 170 noninvasion MRIs, 69 high grade MRIs and 145 low grade MRIs.
For preprocessing, radiologists are asked to crop the tumor ROIs 
following previous works~\cite{adeli2018prediction,joo2021extensive,behling2021brain}.
In order to maintain the shape of tumor and edema area, we zero pad the cropped image into a square and resize them into $128\times128\times24$ as the network input.

\para{Evaluation Metrics and Implementation Details.} 
The metrics used to evaluate the network performance are Sensitivity, Specificity, Accuracy, G-Means, Balanced Accuracy~\cite{brodersen2010balanced}, MCC, AUPRC, and AUC. Also note that MCC and AUPRC are two metrics which can be used with imbalanced datasets.
The proposed algorithm is built with PyTorch on a NVIDIA RTX 3090 GPU. We use the Adam optimizer and set the initial learning rate to $1e-3$. To prevent overfitting, we add dropout of $0.5$ and $L2$ regularization with regularization parameter as $1e-3$. Flip, Gaussian noise and random crop are employed in data augmentation. Our model has a parameter size of 200M, trained with 100 epoches, and takes an average inference time of 0.098s for one image. 

\begin{table}[t]
	\centering
	\caption{Quantitative comparison. The best and second best results for each metric are highlighted in red and blue, respectively.}
	\renewcommand\arraystretch{1.1}
	\label{table_comparison}
	\setlength{\tabcolsep}{1.1mm}{
		\small
		\begin{tabular}{|cc|p{1.2cm}<{\centering}|p{1.2cm}<{\centering}|p{1.2cm}<{\centering}|p{1.2cm}<{\centering}|p{1.2cm}<{\centering}|}
			\hline
			\multicolumn{2}{|c|}{Methods}        & EFMT & MFMT  & MMoE  & MAML  & Proposed \\ \hline
			\multicolumn{1}{|c|}{\multirow{6}{*}{Invasion}} & Sensitivity & \textcolor{blue}{0.7593} & 0.6852 & \textcolor{red}{0.7963} & 0.7407 & \textcolor{blue}{0.7593}    \\ \cline{2-7} 
			\multicolumn{1}{|c|}{}                     & Specificity & 0.9707 & 0.9771 & 0.9630 & \textcolor{red}{0.9824} & \textcolor{blue}{0.9789}    \\ \cline{2-7} 
			\multicolumn{1}{|c|}{}                     & Accuracy & 0.9642 & 0.9681 & 0.9579 & \textcolor{red}{0.9750} & \textcolor{blue}{0.9721}    \\ \cline{2-7} 
			\multicolumn{1}{|c|}{}                     & G-Means  & 0.8555 & 0.8181 & \textcolor{red}{0.8756} & 0.8529 & \textcolor{blue}{0.8619}    \\ \cline{2-7} 
			\multicolumn{1}{|c|}{}                     & Balanced Accuracy  & 0.8650 & 0.8312 & \textcolor{red}{0.8797} & 0.8616 & \textcolor{blue}{0.8691}    \\ \cline{2-7}
			\multicolumn{1}{|c|}{}                     & MCC  & 0.5787 & 0.5691 & 0.5529 & \textcolor{red}{0.6486} & \textcolor{blue}{0.6252}    \\ \cline{2-7}
			\multicolumn{1}{|c|}{}                     & AUPRC  & 0.3607 & 0.3558 & 0.3329 & \textcolor{red}{0.4490} & \textcolor{blue}{0.4157}    \\ \cline{2-7}
			\multicolumn{1}{|c|}{}                     & AUC & 0.9574 & 0.9527 & 0.9425 & \textcolor{blue}{0.9668} & \textcolor{red}{0.9787}    \\ \hline
			
			\multicolumn{1}{|c|}{\multirow{6}{*}{Meningioma}} & Sensitivity & 0.5950  & 0.6582 & 0.6498  & \textcolor{red}{0.6920} & \textcolor{blue}{0.6878}    \\ \cline{2-7} 
			\multicolumn{1}{|c|}{}                     & Specificity & 0.8475 & \textcolor{blue}{0.9132} & 0.8126 & 0.8586 & \textcolor{red}{0.9277}    \\ \cline{2-7} 
			\multicolumn{1}{|c|}{}                     & Accuracy & 0.8134 & \textcolor{blue}{0.8788} & 0.7907 & 0.8362 & \textcolor{red}{0.8953}    \\ \cline{2-7} 
			\multicolumn{1}{|c|}{}                     & G-Means  & 0.7099 & \textcolor{blue}{0.7748} & 0.7262 & 0.7690 & \textcolor{red}{0.7981}    \\ \cline{2-7} 
			\multicolumn{1}{|c|}{}                     & Balanced Accuracy  & 0.6554 & \textcolor{blue}{0.7493} & 0.6460 & 0.6998   & \textcolor{red}{0.7806}    \\ \cline{2-7} 
			\multicolumn{1}{|c|}{}                     & MCC  & 0.3706 & \textcolor{blue}{0.5327} & 0.3671 & 0.4675 & \textcolor{red}{0.5860}    \\ \cline{2-7}
			\multicolumn{1}{|c|}{}                     & AUPRC  & 0.2814 & \textcolor{blue}{0.4088} & 0.2773 & 0.3509 & \textcolor{red}{0.4599}    \\ \cline{2-7}
			\multicolumn{1}{|c|}{}                     & AUC & 0.8137 & \textcolor{blue}{0.8707} & 0.8163 & 0.8618 & \textcolor{red}{0.8870}    \\ \hline
	\end{tabular}}
	\vspace{-0.5cm}
\end{table}

\para{Comparison with Other Methods.}
As there are no existing methods directly available for our joint classification tasks, 
we built four alternative methods for comparison. 
All the compared methods use ResNet34 as the backbone, and are trained on the collected dataset to get best results.
(1) EFMT (Early Fusion with Multi-task). We built a model that concatenates multi-modal MRIs at the input and the extracted features are connected to two classifiers that perform the corresponding tasks.
(2) MFMT (Middle Fusion with Multi-task). The MFMT model contains three ResNet34 to extract the features of multi-modal MRIs respectively, and the concatenated features are fed to two classifiers to finish both tasks. 
(3) MMoE~\cite{ma2018modeling}. We adapt this method to our multi-task topic, which uses three expert networks and two gating networks to generate task-specific features. We adopt ResNet34 as the expert network and ResNet18 as the gating network. 
(4) MAML~\cite{zhang2021modality}. This method extracts features for each modality via multiple encoders, and estimates an modality-aware attention map to obtain boosted features. We change the output of this method to two classifiers. 
It is noted that the classifier adopted in the above compared methods is the same as the auxiliary branch in our proposed framework; see Fig.~\ref{fig:proposed_overview}.

Table~\ref{table_comparison} summarizes the comparison results.
Among the compared methods, for the brain invasion identification task, MMoE gets the highest sensitivity, g-means, and balanced accuracy ($0.7963$, $0.8756$, $0.8797$). MAML achieves the best specificity, accuracy, MCC, AUPRC, and AUC ($0.9824$, $0.9750$, $0.6486$, $0.4490$, $0.9668$). 
For the meningioma grade prediction task, MAML gets the best sensitivity ($0.6920$) and MFMT achieves the best specificity, accuracy, g-means, balanced accuracy, MCC, AUPRC, and AUC ($0.9132$, $0.8788$, $0.7748$, $0.7493$, $0.5327$, $0.4088$, $0.8701$). 
In comparison, our proposed algorithm achieves the best AUC ($0.9787$, $0.0119$ better than the MAML) for the brain invasion identification task; and the best specificity, accuracy, g-means, balanced accuracy, MCC, AUPRC, and AUC ($0.9277$, $0.8953$, $0.7981$, $0.7806$, $0.5860$, $0.4599$, $0.8870$) for the meningioma grade prediction task.
It is noted that for MCC and AUPRC metrics, our method reports the best results for meningioma grade prediction and the second best results for brain invasion identfication.
The standard deviations of metrics results can be found in the supplementary material. We also calculate AUC differences between compared methods and ours using ROC-kit~\cite{metz2004roc}, while p-values are less than 0.05.

\begin{table}[t]
\centering
\caption{The results of ablation experiments. TC, Aux and $L_{con}$ mean Task-common branch, auxiliary branch and contrastive loss respectively. The best and second best results for each metric are highlighted in red and blue, respectively.}
\renewcommand\arraystretch{1.1}
\label{table_ablation}
\setlength{\tabcolsep}{0.9mm}{
\small
\begin{tabular}{|cc|c|c|c|c|c|}
\hline
\multicolumn{2}{|c|}{Baseline}                      & Baseline1 & Baseline2 & Baseline3 & Baseline4 & Proposed \\ \hline
\multicolumn{1}{|c|}{\multirow{3}{*}{Ablation}} & TC     & $\times$     & $\checkmark$     & $\checkmark$     & $\checkmark$     & $\checkmark$     \\ \cline{2-7} 
\multicolumn{1}{|c|}{}                          & $L_{con}$ & $\times$     & $\times$     & $\checkmark$     & $\times$     & $\checkmark$     \\ \cline{2-7} 
\multicolumn{1}{|c|}{}                          & Aux    & $\times$     & $\times$     & $\times$     & $\checkmark$     & $\checkmark$     \\ \hline
\multicolumn{1}{|c|}{\multirow{8}{*}{Invasion}}   & Sensitivity       & 0.6111    & \textcolor{red}{0.7778}    & 0.7407    & 0.7037    & \textcolor{blue}{0.7593}   \\ \cline{2-7} 
\multicolumn{1}{|c|}{}                            & Specificity       & 0.9730    & 0.9736    & 0.9783    & \textcolor{red}{0.9794}    & \textcolor{blue}{0.9789}   \\ \cline{2-7} 
\multicolumn{1}{|c|}{}                            & Accuracy          & 0.9619    & 0.9676    & \textcolor{blue}{0.9710}    & \textcolor{blue}{0.9710}    & \textcolor{red}{0.9721}   \\ \cline{2-7} 
\multicolumn{1}{|c|}{}                            & G-Means           & 0.7629    & \textcolor{red}{0.8699}    & 0.8504    & 0.8291    & \textcolor{blue}{0.8619}   \\ \cline{2-7} 
\multicolumn{1}{|c|}{}                            & Balanced Accuracy & 0.7921    & \textcolor{red}{0.8757}    & 0.8595    & 0.8416    & \textcolor{blue}{0.8691}   \\ \cline{2-7} 
\multicolumn{1}{|c|}{}                            & MCC               & 0.5379    & 0.5990    & \textcolor{blue}{0.6097}    & 0.5974    & \textcolor{red}{0.6252}   \\ \cline{2-7} 
\multicolumn{1}{|c|}{}                            & AUPRC             & 0.3216    & 0.3859    & \textcolor{blue}{0.4026}    & 0.3882    & \textcolor{red}{0.4157}   \\ \cline{2-7} 
\multicolumn{1}{|c|}{}                            & AUC               & 0.9542    & 0.9695    & \textcolor{blue}{0.9727}    & 0.9717    & \textcolor{red}{0.9787}   \\ \hline
\multicolumn{1}{|c|}{\multirow{8}{*}{Meningioma}} & Sensitivity       & 0.5781    & 0.6878    & \textcolor{red}{0.7131}    & \textcolor{blue}{0.6962}    & 0.6878   \\ \cline{2-7} 
\multicolumn{1}{|c|}{}                            & Specificity       & \textcolor{blue}{0.9211}    & 0.8941    & 0.8692    & 0.8988    & \textcolor{red}{0.9277}   \\ \cline{2-7} 
\multicolumn{1}{|c|}{}                            & Accuracy          & \textcolor{blue}{0.8749}    & 0.8663    & 0.8481    & 0.8714    & \textcolor{red}{0.8953}   \\ \cline{2-7} 
\multicolumn{1}{|c|}{}                            & G-Means           & 0.7240    & 0.7841    & 0.7859    & \textcolor{blue}{0.7909}    & \textcolor{red}{0.7981}   \\ \cline{2-7} 
\multicolumn{1}{|c|}{}                            & Balanced Accuracy & \textcolor{blue}{0.7486}    & 0.7289    & 0.7070    & 0.7336    & \textcolor{red}{0.7806}   \\ \cline{2-7} 
\multicolumn{1}{|c|}{}                            & MCC               & 0.4934    & 0.5157    & 0.4903    & \textcolor{blue}{0.5272}    & \textcolor{red}{0.5860}   \\ \cline{2-7} 
\multicolumn{1}{|c|}{}                            & AUPRC             & 0.3728    & 0.3935    & 0.3676    & \textcolor{blue}{0.4016}    & \textcolor{red}{0.4599}   \\ \cline{2-7} 
\multicolumn{1}{|c|}{}                            & AUC               & 0.8851    & 0.8753    & \textcolor{blue}{0.8853}    & 0.8815    & \textcolor{red}{0.8870}   \\ \hline
\end{tabular}}
\vspace{-0.5cm}
\end{table}

\para{Ablation Analysis.}
To demonstrate the effectiveness of the TC (Task-common branch), $Aux$ (auxiliary branch) and $L_{con}$ (contrastive learning loss), we conduct ablation studies; see Table~\ref{table_ablation}.
Compared with the proposed algorithm, Baseline1 removes the task-common branch, auxiliary branch and contrastive learning loss. 
Baseline2 adds the task-common branch, 
while Baseline3 and Baseline4 add auxiliary branch and contrastive learning loss, respectively.

It can be seen from the first two columns that MCC and AUPRC increase from ($0.5379$, $0.3216$) to ($0.5590$, $0.3859$) for invasion task and from ($0.4934$, $0.3728$) to ($0.5157$, $0.3935$) for meningioma task by adding $TC$, which proves the rationality of the task-common branch. 
On this basis, $L_{con}$ is added and improves MCC and AUPRC to ($0.6097$,$0.4026$) for invasion task. 
What's more, respecting the Baseline3, the proposed method adds $L_{aux}$ to further improve MCC and AUPRC to ($0.6252$,$0.4157$) for invasion task and ($0.5860$, $0.4599$) for meningioma task, which verifies that the auxiliary branch can help the contrastive learning strategy to play a better role.
In comparison, the sensitivity of meningioma grade is somehow lower, for which we would like to further explore in the future work. 

\section{Conclusion}
Joint prediction of brain invasion and meningioma grade is a novel topic and an urgent clinical need.
As far as we know, there is no study so far that solves both prediction tasks based on brain MRIs simultaneously. 
In this paper, we propose a novel contrastive learning-based multi-task algorithm, which respects the coherence between tasks and also enhances the distinctness of each invididual task.
We first use a middle-fusion strategy to fuse the feature maps from the multi-modal MRIs. 
Then, we disentangle the fused image features into task-specific features, which focus on a separate task, and task-common features that can perform both tasks. 
A new contrastive loss is leveraged to use task-common features as guidance to improve the prediction ability of two tasks.
In the future work, we would like to explore the influence of each modality MRI on the two tasks to further improving the results of both tasks.

\subsubsection{Acknowledgements} This work was supported by the grant from Tianjin Natural Science Foundation (Grant No. 20JCYBJC00960) and HKU Seed Fund for Basic Research (Project No. 202111159073).


%
%

%
%
%
\bibliographystyle{paper1557}
\bibliography{paper1557}

\begin{thebibliography}{10}
\providecommand{\url}[1]{\texttt{#1}}
\providecommand{\urlprefix}{URL }
\providecommand{\doi}[1]{https://doi.org/#1}

\bibitem{adeli2018prediction}
Adeli, A., Hess, K., Mawrin, C., Streckert, E.M.S., Stummer, W., Paulus, W.,
  Kemmling, A., Holling, M., Heindel, W., Schmidt, R., et~al.: Prediction of
  brain invasion in patients with meningiomas using preoperative magnetic
  resonance imaging. Oncotarget  \textbf{9}(89),  35974 (2018)

\bibitem{behling2021brain}
Behling, F., Hempel, J.M., Schittenhelm, J.: Brain invasion in meningioma—a
  prognostic potential worth exploring. Cancers  \textbf{13}(13), ~3259 (2021)

\bibitem{brodersen2010balanced}
Brodersen, K.H., Ong, C.S., Stephan, K.E., Buhmann, J.M.: The balanced accuracy
  and its posterior distribution. In: 2010 20th international conference on
  pattern recognition. pp. 3121--3124. IEEE (2010)

\bibitem{brokinkel2017brain}
Brokinkel, B., Hess, K., Mawrin, C.: Brain invasion in meningiomas—clinical
  considerations and impact of neuropathological evaluation: a systematic
  review. Neuro-Oncology  \textbf{19}(10),  1298--1307 (2017)

\bibitem{Champeaux2017Recurrence}
Champeaux, C., Houston, D., Dunn, L., meningioma, A.: A study on recurrence and
  disease-specific survival. Neurochirurgie  \textbf{63},  272--281 (2017)

\bibitem{chartsias2019disentangled}
Chartsias, A., Joyce, T., Papanastasiou, G., Semple, S., Williams, M., Newby,
  D.E., Dharmakumar, R., Tsaftaris, S.A.: Disentangled representation learning
  in cardiac image analysis. Medical image analysis  \textbf{58},  101535
  (2019)

\bibitem{chartsias2020disentangle}
Chartsias, A., Papanastasiou, G., Wang, C., Semple, S., Newby, D.E.,
  Dharmakumar, R., Tsaftaris, S.A.: Disentangle, align and fuse for multimodal
  and semi-supervised image segmentation. IEEE transactions on medical imaging
  \textbf{40}(3),  781--792 (2020)

\bibitem{Chen2018multitask}
Chen, Z., Badrinarayanan, V., Lee, C.Y., Rabinovich, A.: Gradnorm: Gradient
  normalization for adaptive loss balancing in deep multitask networks. In:
  International Conference on Machine Learning. pp. 794--803. PMLR (2018)

\bibitem{Doersch2017multitask}
Doersch, C., Zisserman, A.: Multi-task supervised visual learning. In:
  Proceedings of the IEEE/CVF International Conference on Computer Vision
  (2017)

\bibitem{hale2018machine}
Hale, A.T., Stonko, D.P., Wang, L., Strother, M.K., Chambless, L.B.: Machine
  learning analyses can differentiate meningioma grade by features on magnetic
  resonance imaging. Neurosurgical focus  \textbf{45}(5), ~E4 (2018)

\bibitem{han2021meningiomas}
Han, Y., Wang, T., Wu, P., Zhang, H., Chen, H., Yang, C.: Meningiomas:
  preoperative predictive histopathological grading based on radiomics of mri.
  Magnetic Resonance Imaging  \textbf{77},  36--43 (2021)

\bibitem{henaff2020data}
Henaff, O.: Data-efficient image recognition with contrastive predictive
  coding. In: International Conference on Machine Learning. pp. 4182--4192.
  PMLR (2020)

\bibitem{huang2019imaging}
Huang, R.Y., Bi, W.L., Griffith, B., Kaufmann, T.J., la~Foug{\`e}re, C.,
  Schmidt, N.O., Tonn, J.C., Vogelbaum, M.A., Wen, P.Y., Aldape, K., et~al.:
  Imaging and diagnostic advances for intracranial meningiomas. Neuro-oncology
  \textbf{21}(Supplement\_1),  i44--i61 (2019)

\bibitem{joo2021extensive}
Joo, L., Park, J.E., Park, S.Y., Nam, S.J., Kim, Y.H., Kim, J.H., Kim, H.S.:
  Extensive peritumoral edema and brain-to-tumor interface mri features enable
  prediction of brain invasion in meningioma: Development and validation.
  Neuro-oncology  \textbf{23}(2),  324--333 (2021)

\bibitem{KANDEMIRLI2020106205}
Kandemirli, S.G., Chopra, S., Priya, S., Ward, C., Locke, T., Soni, N.,
  Srivastava, S., Jones, K., Bathla, G.: Presurgical detection of brain
  invasion status in meningiomas based on first-order histogram based texture
  analysis of contrast enhanced imaging. Clinical Neurology and Neurosurgery
  \textbf{198},  106205 (2020)

\bibitem{Kendall2018multitask}
Kendall, A., Gal, Y., Cipolla, R.: Multi-task learning using uncertainty to
  weigh losses for scene geometry and semantics. In: Proceedings of the IEEE
  conference on computer vision and pattern recognition. pp. 7482--7491 (2018)

\bibitem{li2021clinical}
Li, N., Mo, Y., Huang, C., Han, K., He, M., Wang, X., Wen, J., Yang, S., Wu,
  H., Dong, F., et~al.: A clinical semantic and radiomics nomogram for
  predicting brain invasion in who grade ii meningioma based on tumor and
  tumor-to-brain interface features. Frontiers in oncology p.~4362 (2021)

\bibitem{louis20072007}
Louis, D.N., Ohgaki, H., Wiestler, O.D., Cavenee, W.K., Burger, P.C., Jouvet,
  A., Scheithauer, B.W., Kleihues, P.: The 2007 who classification of tumours
  of the central nervous system. Acta neuropathologica  \textbf{114}(2),
  97--109 (2007)

\bibitem{louis20162016}
Louis, D.N., Perry, A., Reifenberger, G., Von~Deimling, A., Figarella-Branger,
  D., Cavenee, W.K., Ohgaki, H., Wiestler, O.D., Kleihues, P., Ellison, D.W.:
  The 2016 world health organization classification of tumors of the central
  nervous system: a summary. Acta neuropathologica  \textbf{131}(6),  803--820
  (2016)

\bibitem{ma2018modeling}
Ma, J., Zhao, Z., Yi, X., Chen, J., Hong, L., Chi, E.H.: Modeling task
  relationships in multi-task learning with multi-gate mixture-of-experts. In:
  Proceedings of the 24th ACM SIGKDD International Conference on Knowledge
  Discovery \& Data Mining. pp. 1930--1939 (2018)

\bibitem{nowosielski2017diagnostic}
Nowosielski, M., Galldiks, N., Iglseder, S., Kickingereder, P., Von~Deimling,
  A., Bendszus, M., Wick, W., Sahm, F.: Diagnostic challenges in meningioma.
  Neuro-oncology  \textbf{19}(12),  1588--1598 (2017)

\bibitem{Ostrom2020CBTRUS}
Ostrom, Q.T., Patil, N., Cioffi, G., Waite, K., Kruchko, C., Barnholtz-Sloan,
  J.S.: Cbtrus statistical report: Primary brain and other central nervous
  system tumors diagnosed in the united states in 2013–2017. Neuro Oncology
  \textbf{22},  iv1--iv96 (2020)

\bibitem{park2019radiomics}
Park, Y.W., Oh, J., You, S.C., Han, K., Ahn, S.S., Choi, Y.S., Chang, J.H.,
  Kim, S.H., Lee, S.K.: Radiomics and machine learning may accurately predict
  the grade and histological subtype in meningiomas using conventional and
  diffusion tensor imaging. European radiology  \textbf{29}(8),  4068--4076
  (2019)

\bibitem{metz2004roc}
Pesce~LL, Papaioannu~J, M.C.: Roc-kit software (2004),
  http://radiology.uchicago.edu/?q=MetzROCsoftware

\bibitem{wang2021exploring}
Wang, W., Zhou, T., Yu, F., Dai, J., Konukoglu, E., Van~Gool, L.: Exploring
  cross-image pixel contrast for semantic segmentation. In: Proceedings of the
  IEEE/CVF International Conference on Computer Vision. pp. 7303--7313 (2021)

\bibitem{xie2020delving}
Xie, J., Zhan, X., Liu, Z., Ong, Y.S., Loy, C.C.: Delving into inter-image
  invariance for unsupervised visual representations. arXiv preprint
  arXiv:2008.11702  (2020)

\bibitem{xie2021propagate}
Xie, Z., Lin, Y., Zhang, Z., Cao, Y., Lin, S., Hu, H.: Propagate yourself:
  Exploring pixel-level consistency for unsupervised visual representation
  learning. In: Proceedings of the IEEE/CVF Conference on Computer Vision and
  Pattern Recognition. pp. 16684--16693 (2021)

\bibitem{zhang2021deep}
Zhang, H., Mo, J., Jiang, H., Li, Z., Hu, W., Zhang, C., Wang, Y., Wang, X.,
  Liu, C., Zhao, B., et~al.: Deep learning model for the automated detection
  and histopathological prediction of meningioma. Neuroinformatics
  \textbf{19}(3),  393--402 (2021)

\bibitem{zhang2020radiomics}
Zhang, J., Yao, K., Liu, P., Liu, Z., Han, T., Zhao, Z., Cao, Y., Zhang, G.,
  Zhang, J., Tian, J., et~al.: A radiomics model for preoperative prediction of
  brain invasion in meningioma non-invasively based on mri: A multicentre
  study. EBioMedicine  \textbf{58},  102933 (2020)

\bibitem{zhang2021modality}
Zhang, Y., Yang, J., Tian, J., Shi, Z., Zhong, C., Zhang, Y., He, Z.:
  Modality-aware mutual learning for multi-modal medical image segmentation.
  In: International Conference on Medical Image Computing and Computer-Assisted
  Intervention. pp. 589--599. Springer (2021)

\bibitem{zhu2019deep}
Zhu, Y., Man, C., Gong, L., Dong, D., Yu, X., Wang, S., Fang, M., Wang, S.,
  Fang, X., Chen, X., et~al.: A deep learning radiomics model for preoperative
  grading in meningioma. European journal of radiology  \textbf{116},  128--134
  (2019)

\end{thebibliography}
%




\end{document}